\documentclass[runningheads]{llncs}
\usepackage[T1]{fontenc}
\usepackage{graphicx,verbatim}
\usepackage{booktabs}
\usepackage{amssymb}
\usepackage{makecell}
\usepackage{hyperref} 
\usepackage{multirow}
\usepackage[table]{xcolor}
\usepackage{color}

\begin{document}
\title{Whole-body Representation Learning For Competing Preclinical Disease Risk Assessment}
\titlerunning{Whole-body Representation Learning for Competing Risk Assessment}
%
%
\author{Dmitrii Seletkov\inst{1, 2}\orcidID{0009-0003-0568-1132} \and
Sophie Starck\inst{2}\orcidID{0000-0003-2495-6114} \and
Ayhan Can Erdur\inst{2}\orcidID{0009-0006-0290-0434} \and Yundi Zhang\inst{2}\orcidID{0009-0008-7725-6369} \and Daniel Rueckert\inst{2,3}\orcidID{0000-0002-5683-5889} \and Rickmer Braren \inst{1,4}\orcidID{0000-0001-6039-6957}}

%
\authorrunning{Seletkov et al.}
%
\institute{Institute of Diagnostic and Interventional Radiology, Technical University of Munich, School of Medicine, Munich, Germany \and Chair for AI in Healthcare and Medicine, Technical University of Munich (TUM) and TUM University Hospital, Munich, Germany \and Department of Computing, Imperial College London, London, UK \and German Cancer Consortium (DKTK), Munich partner site, Heidelberg, Germany \\
\email{dmitrii.seletkov@tum.de}}
\maketitle              
\begin{abstract}
Reliable preclinical disease risk assessment is essential to move public healthcare from reactive treatment to proactive identification and prevention. However, image-based risk prediction algorithms often consider one condition at a time and depend on hand-crafted features obtained through segmentation tools. We propose a whole-body self-supervised representation learning method for the preclinical disease risk assessment under a competing risk modeling. This approach outperforms whole-body radiomics in multiple diseases, including cardiovascular disease (CVD), type 2 diabetes (T2D), chronic obstructive pulmonary disease (COPD), and chronic kidney disease (CKD). Simulating a preclinical screening scenario and subsequently combining with cardiac MRI, it sharpens further the prediction for CVD subgroups: ischemic heart disease (IHD), hypertensive diseases (HD), and stroke. The results indicate the translational potential of whole-body representations as a standalone screening modality and as part of a multi-modal framework within clinical workflows for early personalized risk stratification. The code is available at~\url{https://github.com/yayapa/WBRLforCR/}


\end{abstract}
\section{Introduction}
Cardiometabolic, oncologic, neurodegenerative, and inflammatory diseases collectively not only claim over 26 million lives each year, but reduce healthy life expectancy and, thus, present a massive economic burden across the globe~\cite{Naghavi24}. Personalized risk assessment and early disease detection enable shifting the focus towards prevention. In this context, advances in medical imaging offer an opportunity to integrate imaging into clinical prediction and detection processes. 

In the past decades, the primary focus of risk assessment has been on CVD as a major cause of reduced quality of life and early death ~\cite{who_cvd_2021}. Recent works address CVD risk prediction using scalable digital tools~\cite{Dolezalova21CVD}, clinical data~\cite{Steinfeldt22}, and multifactorial analyses incorporating behavioral, socioeconomic, and environmental variables~\cite{Mamouei23}. Other conditions have also been explored through non-image data. ~\cite{Julkunen23} evaluates metabolic blood biomarkers for CKD, while for cancer, ~\cite{Sun23} examines liver function biomarkers, ~\cite{Soto22} and~\cite{Chang23} dietary factors.

However, the use of imaging data for risk assessment remains underexplored. ~\cite{Prasad24} extracts vascular patterns from retinal fundus images for CVD prognosis. ~\cite{Huang24} improves T2D risk prediction based on genetic data using image-derived features from abdominal ultrasonography and bone mineral density scans. ~\cite{Linge23} uses whole-body and liver MRIs to extract muscle composition and liver biomarkers to evaluate all-cause mortality risk.

Despite a very high frequency of comorbidities, prognostic algorithms~\cite{Prasad24,Huang24,Lian23} commonly address single risks, ignoring competing disease modeling, which may undermine interplay and precision in risk assessment~\cite{Jeanselme23,Wiegrebe24}. Furthermore, in the case of image data, they rely on segmentation tools to extract features, reducing the higher dimensions of the images important for the training of the deep time-to-event risk models~\cite{Haarburger19}. Self-supervised representation learning (RL) may remove this bottleneck, enabling learning informative embeddings without annotation. This is well-explored for cardiac MRI~\cite{Biffi18,Sun22,Zhang24} and is largely untapped in whole-body MRI, despite its clinical ability to provide a multi-organ view in a single screening exam.

\begin{figure}[!ht]
    \centering
    \includegraphics[width=\textwidth]{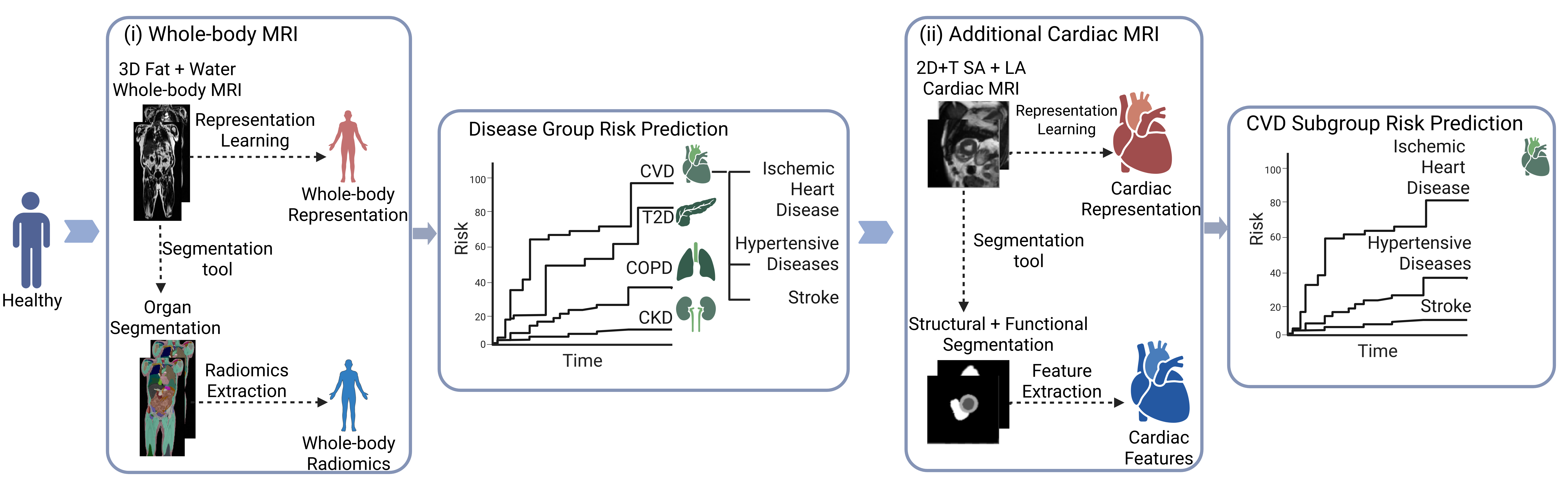}
    \caption{The proposed clinical workflow using whole-body MRI screening includes (i) competing risk assessment of the disease groups: CVD, T2D, COPD, and CKD; (ii) CVD subgroups: IHD, HD, and stroke. (ii) incorporates the additional cardiac MRI. Created in BioRender. Seletkov, D. (2025)~\url{https://BioRender.com/
swjgnva}.} 
    \label{fig:study_design}
\end{figure}

In this work, we first show how the prospective study UK Biobank~\cite{Sudlow15} can be used to investigate competing preclinical disease risk assessment using the radiation-free whole-body MRI screening exams for the multiple disease groups, i.e., CVD, T2D, COPD, and CKD. We then investigate an image-based workflow for further stratification of one disease group (CVD) into subgroups (ischemic heart disease (IHD), hypertensive diseases (HD), and stroke) by adding an organ-specific (cardiac) MRI. We introduce a self-supervised whole-body representation learning method for competing preclinical disease risk assessment. These whole-body representations are evaluated against whole-body radiomics, cardiac structural and functional features, and cardiac representations for the disease groups and CVD subgroups. We demonstrate that whole-body representations can be effectively used standalone for the preclinical disease risk assessment, outperforming the other single modalities, and also in combination with cardiac features, as a multi-modal image-based risk assessment tool for CVD subgroups. The overview of our work is presented in Fig.~\ref{fig:study_design}. 

\section{Method}
\subsection{Datasets Construction}
UK Biobank~\cite{Sudlow15} is a long-term population study monitoring 500,000 volunteers 40–69 years of age at recruitment in 2006–2010. It includes linkage to electronic health‑related records, such as hospital admissions, and self‑reported data collected via interview, enabling disease identification. 100,000 participants are recalled for an imaging assessment, including whole-body and cardiac MRIs.

The whole-body MRI consists of a neck-to-knee T1-weighted dual-echo Dixon MR image center-cropped to a size of $[220\times160\times360]$ and a resolution of $[2.23\times3\times2.23]$ mm with water and fat contrasts, resulting in a tensor of shape $(220, 160, 360, 2)$. Cardiac MR imaging captures dynamic 2D sequences from multiple anatomical views across the cardiac cycle. Each subject scan includes 6 short-axis and 3 long-axis slices, center-cropped to $128\times128$ pixels, acquired over $50$ time frames, forming a tensor of shape $(128, 128, 9, 50)$. 

We construct the datasets for the competing preclinical disease risk assessment using UK Biobank as follows.
We identify the disease groups (T2D, COPD, CKD, and CVD) and CVD subgroups (IHD, HD, and stroke), using linked health sources, including hospital in-patient summary diagnoses (FIDs 41270, 41280, 41271, 41281) and self-reported diseases (FIDs 20002, 20008). For each disease, a set of ICD-10, ICD-9, and self-reported codes is specified by a medical expert on the basis provided by~\cite{Quan05,Tian23} and used as inclusion criteria. 

An event is defined as the recorded occurrence of a disease-specific diagnosis in any of the available linked health sources. A subject is assigned to a disease group or subgroup if the first occurrence of any disease-specific event is recorded, and no relevant event occurs before or within three months after imaging. The three-month exclusion period is chosen to minimize the risk of imminent, yet unreported, diagnoses at the time of imaging. The time-to-event label corresponds to the time from imaging until the first occurrence of the disease. Importantly, subjects with a first event from one disease group do not have prior events from any other groups. As a healthy control group, we select subjects without any existing event at any time before or after the imaging. The time-to-event label corresponds to the time from the imaging until the censoring date provided by the UK Biobank. 

The disease group dataset contains the CVD, T2D, COPD, CKD, and healthy (censored) subjects. The CVD subgroup dataset contains the CVD subjects divided into the IHD, HD, and stroke.
Table~\ref{tab:datasets} overviews the constructed datasets.

\begin{table}[!ht]
  \centering
  \caption{Overview of the number of events and average time-to-event ($\overline{TTE}$) with SD in years across competing risks for disease groups and CVD subgroups datasets.}
  \label{tab:datasets}
  \begin{tabular}{llccccc c cccc}
    \toprule
    & & \multicolumn{5}{c}{Disease groups} & & \multicolumn{4}{c}{CVD subgroups} \\
    \cmidrule(lr){3-7} \cmidrule(lr){9-12}
     & & CVD & T2D & COPD & CKD & Healthy & & IHD & HD & Stroke & Healthy \\
    \midrule
    \# events &  & 1536 & 93 & 106 & 147 & 1139 & & 382 & 1044 & 110 & 1139 \\
    \midrule
    $\overline{TTE}$ (years)&  & 
    \makecell{3.0{\scriptsize (1.8)}} & 
    \makecell{2.9{\scriptsize (1.7)}} & 
    \makecell{3.2{\scriptsize (2.0)}} & 
    \makecell{3.0{\scriptsize (1.8)}} & 
    \makecell{4.1{\scriptsize (1.6)}} & &
    \makecell{2.8{\scriptsize (1.8)}} & 
    \makecell{3.1{\scriptsize (1.8)}} & 
    \makecell{2.9{\scriptsize (1.8)}} & 
    \makecell{4.1{\scriptsize (1.6)}} \\
    \bottomrule
  \end{tabular}
\end{table}

\subsection{Whole-body Radiomics and Cardiac Features}
For the cardiac MRI, we follow the approach by Bai et al.~\cite {Bai20}, which includes applying a segmentation tool~\cite{Bai20} and extracting cardiac structural and functional features. These features contain volumetric measurements of all four cardiac chambers, including end-systolic and end-diastolic volumes. Functional features contain volumetric measurements, such as ejection fractions, stroke volume, and ventricular cardiac output. Structural measurements include the ventricular mass and the detailed assessment of myocardial wall thickness. More details on these features and their clinical relevance can be found in the original work~\cite{Bai20}.
 
For the whole-body MRI, we apply a whole-body MRI segmentation tool~\cite{Graf24} and extract radiomics features using PyRadiomics~\cite{pyradiomics}. 
The radiomics include first-order statistics, gray level co-occurrence matrix, gray level run length matrix, gray level size zone matrix, neighboring gray-tone difference matrix, gray level dependence matrix, and shape-based features. Except for shape-based features, we extract radiomics features for fat and water contrasts separately. To reduce the dimensionality but represent different organ systems, we split the radiomics features into anatomical categories: heart, vascular, respiratory, digestive, liver, pancreas, spleen, urinary,  endocrine, kidney, spine, bone, muscle, and fat. For each category, we apply PCA dimensionality reduction to extract $10$ features per category. This reduces the number of features from 14,000 to 140.

\subsection{Representation Learning}
To extract cardiac representations, we follow the approach of Zhang et al.~\cite{Zhang24}, based on a Masked Autoencoder (MAE)~\cite{He22} to train a model on $14,000$ cardiac MRIs. This consists of $6$ encoder layers and $2$ decoder layers with an encoder embedding dimension of $1025$. The patch size is $[8\times8\times25]$ with $8$ the spatial and $25$ the temporal dimension. The mask ratio is set to $70\%$. We use Adam optimizer with a constant learning rate $10^{-4}$ and a weight decay of $0.05$. 

To extract whole-body representations, we develop a model based on Masked Autoencoders (MAE)~\cite{He22}, similar to~\cite{Zhang24}. For each contrast, we decompose each 3D volume into 3D patches. For localization, each patch is supplied with a 4D positional embedding~\cite{He22}, indicating the spatial index and the corresponding contrast. Using intensities, we separate the background from the whole-body parts in the MRI. We mask out randomly $70\%$ of all foreground patches. The remaining patches, $X'$, are fed into an encoder $E$ to learn a whole-body representation. A lightweight decoder $D$ is applied to predict the masked patches and reconstruct the original image $X$. The objective is to minimize the mean squared error (MSE) loss function between the original image and the reconstructed one $L = ||X-D(E(X'))||$. Compared to the cardiac RL model, we use the patch size $[15\times10\times10]$. We use the same architecture and the training strategy as for cardiac RL to train a model on the 70,000 whole-body MRIs, representing the general population, regardless of their disease status.

\subsection{Competing Risk}
We model our data as a set of $\{x_i, t_i, e_i\}$, with $x_i\in\mathbb{R}^d$ the $d$-dimensional input, $t_i \in \mathbb{R^+}$ the time of the first disease occurrence, $e_i \in [0, R]$ the disease for the subject $i$, and $R$ a number of competing risks, where $0$ indicates right censoring. Competing risk modeling estimates the Cumulative Incidence Function (CIF), $F_r$, for each disease $r$, namely the probability of observing the disease $r$ before time $t$ without prior occurrence of any other competing diseases, i.e., $F_r(t|x) = \mathbb{P}(T < t, risk=r|x)$. 

We consider three state-of-the-art deep competing risk models: Deep Survival Machines (DSM)~\cite{Nagpal21}, Neural Fine Gray (NFG)~\cite{Jeanselme23}, and DeepHit~\cite{Lee18}.

DSM model the CIF as a mixture of $k$ primitive distributions (Weibull or Log-normal), $G(t, x; \xi, \varsigma, \beta, \eta)$, parameterized by the neural network:

\begin{equation}
    F_r(t|x) = \sum^k_{j=1}\pi_{r,j}(E(x)) \cdot G_{r,j}(t; E(x))
\end{equation}

where $E$ is an encoder network and $\xi, \varsigma, \beta, \eta$ are learned parameters.

NFG models the CIF through the monotonic neural network architecture:

\begin{equation}
F_r(t|x) = B(E(x))_r \cdot (1 - exp(-t \times M_r(t, E(x))))
\end{equation}
where $B$ is an MLP with a final $softmax$ function to balance the probability of observing each risk; $M_r$ is a positive monotonic neural network constrained to have its outcome positive and monotonic, representing the risk distribution.

DeepHit models the joint distribution of competing risks by discretizing the output times into $L$ segments using risk-specific subnetworks. Hence, the model predicts $y \in \mathbb{R}^{L \times R}$, where the prediction for each time segment $t_l$: 
\begin{equation}
    F_r(t_l|x) = y_{t_l} = \sum^l_{j=1}softmax(concat^R_{q=1}(H_q(E(x), x)))
\end{equation}

where $H_q$ is a risk-specific subnetwork and $E$ is a shared network. 

We use all the above approaches to estimate the CIF for the competing preclinical disease risk assessment. For whole-body radiomics and cardiac features experiments, MRI scans are processed through segmentation and feature extraction tools. For the learned representations, we use the corresponding RL encoders. Each set of features is then fed into a competing risk model.

\subsection{Metrics and Evaluation}
As a primary evaluation metric, we use the time-dependent concordance index $C^{td}$-index~\cite{Antolini05}. This assesses the ability of the model to order the relative risks pairwise, i.e., $ \mathbb{P}(F_r(t|x_i) > F_r(t|x_j)|e_i=r, t_i< t_j, t_i \leq t)$. 

Following the evaluation protocol of Jeanselme et al.~\cite{Jeanselme23}, we apply a disease-stratified, nested 5-fold cross-validation with $10\%$ of each training set left out for hyperparameter tuning. The reported results are the means and t-score 95\% CI across five test folds. Random search is used on the following grid over 100 iterations for all models. Learning rate ($[10^{-2}, 10^{-4}]$), batch size ($[100, 1000]$), dropout ($\{0, 0.25, 0.5, 0.75\}$), number of layers ($[1, 4]$) and nodes ($\{32, 64, 128, \\256, 512\}$). All models are optimized using an Adam optimizer over 1,000 epochs, with an early stopping criterion computed on a $10\%$ left-out training subset. For DSM, we explore $\{Log-Normal, Weibull\}$ distributions and use 10,000 warm-up iterations ~\cite{Nagpal21}. For DeepHit, we use a $L = 15$ split time discretization.

\section{Results \& Discussion}

\subsection{Representation Learning Evaluation} 
\begin{figure}[t]
\centering
\includegraphics[width=\textwidth]{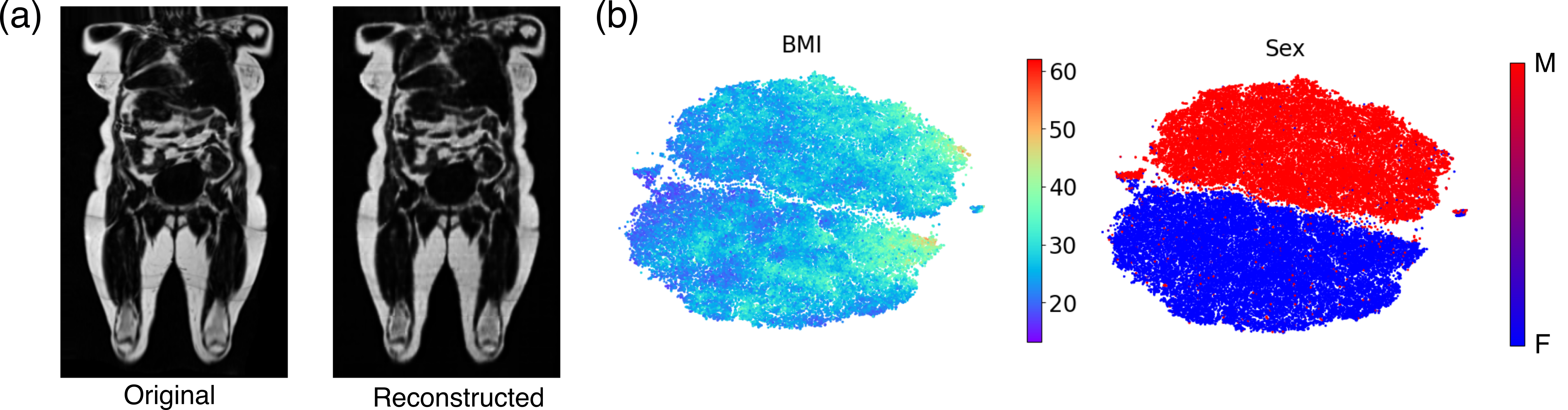}
\caption{Representation Learning evaluation: (a) reconstruction on a random sample (PSNR=37) (b) t-SNE visualization of whole-body representations for BMI (left) and sex (right) classes.} \label{fig:recon_qualitative}
\end{figure}

We evaluate the RL model on the reconstruction, where a better quality indicates a stronger ability of the model to capture the underlying data structure. An example of the qualitative performance is illustrated in Fig.~\ref{fig:recon_qualitative} (a). Quantitatively, our model achieves the mean PSNR of \textbf{32.18} on the 100 held-out samples. Furthermore, we provide the t-SNE visualization of the whole-body representations in Fig.~\ref{fig:recon_qualitative} (b). For this, we label the projected representations with sex and BMI classes. Notably, even without using any labels during RL, the clusters in the model's latent space align well with the sex and BMI of the subjects.

\subsection{Disease Groups}
To explore the translational potential of whole-body screening in the clinical workflow, we first examine the predictive ability of whole-body radiomics and representations for the CVD, T2D, COPD, and CKD. For clarity, we compare the DSM model performances across modalities. Furthermore, we point out the superior performance of DSM compared to NFG and DeepHit and attribute this to its parametric assumption and robustness for smaller-scale datasets~\cite{Jeanselme23}.   

Table~\ref{tab:wb_results} shows that the whole-body representations outperform the whole-body radiomics in terms of $C^{td}$-index, although the margin for CVD, T2D, and CKD is comparatively narrow. We associate this with the better ability of the representations to learn holistic cross-organ interactions in the body compared to the localized and pre-defined radiomics features. 

Note that we intentionally use the frozen pretrained whole-body RL encoder without fine-tuning it to enable training with a larger batch size, essential for the ranking loss component of the deep competing risk models~\cite{Haarburger19}, which is infeasible with high-dimensional whole-body 4D images.

\subsection{CVD Subgroups}
A similar pattern emerges in the risk prediction of CVD subgroups, shown in Table~\ref{tab:cardiac_results}. The whole-body representations consistently surpass the radiomics features, reinforcing their advantage in risk stratification. Notably, the whole-body representations outperform the cardiac structural and functional features and cardiac representations for the risk prediction of CVD subgroups. We associate this with the access to a broader anatomical context, capturing known risk factors, e.g., adipose tissue distribution~\cite{Oikonomou18} or other non-cardiac physiological cues. 

As the next step, we assess the added value of the organ-specific cardiac MRI for distinguishing CVD subgroups. We observe that the cardiac features outperform cardiac representations. This may be attributed to the presence of the structural and functional hand-crafted biomarkers in contrast to the whole-body MRI with automatically segmented structural radiomics. However, extracting these features requires an additional segmentation model and a medical expert prior, adding complexity to the workflow. 

Finally, we investigate a multi-modal setting for CVD subgroup risk prediction by fusing the best-performing modalities. Our results show that incorporating the whole-body representations improves the preclinical CVD subgroup risk prediction beyond that of cardiac MRI alone, highlighting the importance of whole-body screenings.

\begin{table}[!ht]
  \centering
  \caption{Comparison of $C^{td}$-index (mean and 95\% CI across 5 folds) between modalities for preclinical disease risk assessment of CVD, T2D, COPD, and CKD.
  \textbf{Bold} indicates the best, and \textit{italics} the second-best performance across modalities for DSM.}
  \label{tab:wb_results}
  \begin{tabular}{lllcccc}
    \toprule
    \multicolumn{3}{c}{} &  \multicolumn{4}{c}{ $C^{td}$-index${^\uparrow}$} \\
    \cmidrule(lr){4-7}
    Modality & Model &  & CVD & T2D & COPD & CKD \\
    \midrule
        \rowcolor{gray!20} 
    \multirow{3}{*}{\cellcolor{white} \makecell[{{l}}]{Whole \\ body \\ Rad.}}

    & DSM & 
    & \makecell{\textit{0.614}{\tiny (0.599, 0.628)}} & \makecell{\textit{0.692}{\tiny (0.652, 0.733)}} & \makecell{\textit{0.638}{\tiny (0.547, 0.730)}} & \makecell{\textit{0.610}{\tiny (0.538, 0.682)}} \\
    & DeepHit & 
    & \makecell{0.576{\tiny (0.557, 0.595)}} & \makecell{0.531{\tiny (0.414, 0.648)}} & \makecell{0.424{\tiny (0.340, 0.508)}} & \makecell{0.511{\tiny (0.437, 0.586)}} \\
    & NFG & 
    & \makecell{0.627{\tiny (0.617, 0.637)}} & \makecell{0.517{\tiny (0.309, 0.726)}} & \makecell{0.497{\tiny (0.262, 0.732)}} & \makecell{0.507{\tiny (0.349, 0.664)}} \\
    \midrule
        \rowcolor{gray!20} 
    \multirow{3}{*}{\cellcolor{white} \makecell[{{l}}]{Whole \\ body \\ Repr.}} 
    & DSM & 
    & \makecell{\textbf{0.628}{\tiny (0.615,0.642)}} & \makecell{\textbf{0.712}{\tiny (0.567,0.857)}} & \makecell{\textbf{0.682}{\tiny (0.608,0.756)}} & \makecell{\textbf{0.636}{\tiny (0.576,0.696)}}\\
    & DeepHit & 
    & \makecell{0.608{\tiny (0.596, 0.620)}} & \makecell{0.607{\tiny (0.512, 0.702)}} & \makecell{0.530{\tiny (0.398, 0.663)}} & \makecell{0.607{\tiny (0.568, 0.645)}} \\
    & NFG & 
    & \makecell{0.613{\tiny (0.563, 0.663)}} & \makecell{0.492{\tiny (0.349, 0.635)}} & \makecell{0.500{\tiny (0.312, 0.688)}} & \makecell{0.494{\tiny (0.340, 0.647)}} \\
    \bottomrule
  \end{tabular}
\end{table}

\begin{table}[!ht]

  \centering
  \caption{Comparison of $C^{td}$-index (mean and 95\% CI across 5 folds) between modalities for preclinical disease risk assessment of CVD subgroups.
  \textbf{Bold} indicates the best performance, and \textit{italics} the second-best performance across modalities for DSM. }
  \label{tab:cardiac_results}
  \begin{tabular}{lllccc}
    \toprule
    & & & \multicolumn{3}{c}{$C^{td}$-index${^\uparrow}$} \\
    \cmidrule(lr){4-6}
    Modality & Model &  & IHD & HD & Stroke \\
    \midrule
        \rowcolor{gray!20} 
    \multirow{3}{*}{\cellcolor{white} \makecell[{{l}}]{Whole \\ body \\ Rad.}}
    & DSM & 
    & \makecell{0.616{\tiny (0.580, 0.651)}} & \makecell{0.632{\tiny (0.585, 0.679)}} & \makecell{0.519{\tiny (0.378, 0.660)}} \\
    & DeepHit &
    & \makecell{0.578{\tiny (0.526, 0.631)}} & \makecell{0.597{\tiny (0.574, 0.621)}} & \makecell{0.454{\tiny (0.404, 0.504)}} \\
    & NFG &
    & \makecell{0.595{\tiny (0.512, 0.677)}} & \makecell{0.595{\tiny (0.500, 0.691)}} & \makecell{0.468{\tiny (0.358, 0.578)}} \\
    \midrule
        \rowcolor{gray!20} 
    \multirow{3}{*}{\cellcolor{white} \makecell[{{l}}]{Whole \\ body \\ Repr.}}
    & DSM &
    & \makecell{\textit{0.642}{\tiny (0.606, 0.677)}} & \makecell{\textit{0.650}{\tiny (0.636, 0.665)}} & \makecell{\textit{0.614}{\tiny (0.600, 0.626)}} \\
    & DeepHit &
    & \makecell{0.652{\tiny (0.592, 0.711)}} & \makecell{0.649{\tiny (0.614, 0.683)}} & \makecell{0.574{\tiny (0.422, 0.726)}} \\
    & NFG &
    & \makecell{0.609{\tiny (0.577, 0.641)}} & \makecell{0.616{\tiny (0.602, 0.629)}} & \makecell{0.522{\tiny (0.441, 0.603)}} \\
    \midrule
    \rowcolor{gray!20} 
    \multirow{3}{*}{\cellcolor{white} \makecell[{{l}}]{Cardiac \\ Features}}
    & DSM &
    & \makecell{0.625{\tiny (0.588, 0.662)}} & \makecell{0.598{\tiny (0.558, 0.637)}} & \makecell{0.532{\tiny (0.431, 0.632)}} \\
    & DeepHit &
    & \makecell{0.612{\tiny (0.553, 0.672)}} & \makecell{0.602{\tiny (0.559, 0.645)}} & \makecell{0.530{\tiny (0.382, 0.679)}} \\
    & NFG &
    & \makecell{0.536{\tiny (0.443, 0.630)}} & \makecell{0.587{\tiny (0.515, 0.659)}} & \makecell{0.485{\tiny (0.463, 0.507)}} \\
    
    \midrule
    \rowcolor{gray!20} 
    \multirow{3}{*}{\cellcolor{white} \makecell[{{l}}]{Cardiac \\ Repr.}}
    & DSM &
    & \makecell{0.577{\tiny (0.535, 0.619)}} & \makecell{0.587{\tiny (0.562, 0.612)}} & \makecell{0.471{\tiny (0.428, 0.513)}} \\
    & DeepHit &
    & \makecell{0.571{\tiny (0.529, 0.613)}} & \makecell{0.574{\tiny (0.536, 0.612)}} & \makecell{0.471{\tiny (0.369, 0.574)}}\\
    & NFG &
    & \makecell{0.544{\tiny (0.475, 0.614)}} & \makecell{0.570{\tiny (0.520, 0.620)}} & \makecell{0.481{\tiny (0.389, 0.573)}} \\
    
    \midrule
        \rowcolor{gray!20} 
    \multirow{3}{*}{\cellcolor{white} \makecell[{{l}}]{Whole-body Repr.\\ ~~~~~~~~~~~+ \\ Cardiac Features}}
    & DSM &
    & \makecell{\textbf{0.672}{\tiny (0.636, 0.708)}} & \makecell{\textbf{0.665}{\tiny (0.649, 0.682)}} & \makecell{\textbf{0.617}{\tiny (0.486, 0.749)}} \\
    & DeepHit &
    & \makecell{0.632{\tiny (0.577, 0.687)}} & \makecell{0.637{\tiny (0.618, 0.657)}} & \makecell{0.467{\tiny (0.424, 0.511)}} \\
    & NFG &
    & \makecell{0.667{\tiny (0.624, 0.710)}} & \makecell{0.680{\tiny (0.644, 0.716)}} & \makecell{0.555{\tiny (0.404, 0.707)}} \\

    \bottomrule
  \end{tabular}
\end{table}

\section{Conclusion}
This work proposes a whole-body representation learning method for whole-body MRI, showcasing its translational potential within the clinical workflow for preclinical disease risk assessment under competing risk modeling. We demonstrate the utility and robustness of the learned representations for multiple disease groups and CVD subgroups as a standalone modality and in multi-modal settings when combined with image-derived cardiac features. 

The results and generalizability of our experiments should be viewed under the following limitations. Disease group selection is based on the absence of reported diagnoses at imaging and within the subsequent 3 months, without radiological validation by medical experts. Our work focuses on evaluating the whole-body imaging and does not include the clinical information. Future work may prioritize the fusion of image and non-image data to enable more comprehensive risk modeling. Furthermore, our work may be extended to the analysis of the other organ-specific MRIs, such as those of the liver and pancreas,  and their multi-modal fusion. Finally, future work may investigate additional applications of the whole-body RL in diagnostics, phenotyping, or biomarker discovery. 

\textbf{Prospect of application.} Motivated by clinical workflow, our method paves the way towards radiation-free MRI-based population screening, offering early risk stratification and prevention for multiple conditions.

\begin{credits}
\subsubsection{\ackname} For this study, permission to access and analyze the UK Biobank data was approved under the application 87802. This work benefited from resources provided by the joint project "Open Medical Inference," a Module 3 project of the Medical Informatics Initiative of the Federal Government, funded by the German Federal Ministry for Research, Technology, and Space (grant number 01ZZ2315B). Additionally, this work is funded by the European Research Council project Deep4MI (884622).

\subsubsection{\discintname}
The authors declare no competing interests.  
\end{credits}
%
%
%
%
\bibliographystyle{splncs04}
\bibliography{Paper-44}
\end{document}